\documentclass[conference]{IEEEtran}

\ifCLASSINFOpdf
  \usepackage[pdftex]{graphicx}
\fi

\usepackage[cmex10]{amsmath}
\usepackage{amsthm}
\usepackage{amssymb}
\usepackage[utf8]{inputenc}
\usepackage{color}

\usepackage{algorithm}
\usepackage{algcompatible}
\usepackage{booktabs}
\usepackage{float}

\newtheorem{mydef}{Definition}

\begin{document}

\title{Experimental Parameter Analysis\\of the Reservoirs Observers\\ using Echo State Network Approach}

\author{\IEEEauthorblockN{Diana C. Roca Arroyo, Josimar E. Chire Saire}
\IEEEauthorblockA{Institute of Mathematics and Computer Science (ICMC) \\
University of São Paulo (USP)\\
 São Carlos, SP, Brazil\\
 dianaroca@usp.br, jecs89@usp.br}
 }




%


\maketitle

\begin{abstract}
Dynamical systems has a variety of applications for the new information generated during the time. Many phenomenons like physical, chemical or social are not static, then an analysis over the time is necessary. In this work, an experimental analysis of parameters of the model Echo State Network is performed and the influence of the kind of Complex Network is explored to understand the influence on the performance. The experiments are performed using the Rossler attractor.

\end{abstract}


\IEEEpeerreviewmaketitle

\section{INTRODUCTION}

Frequently, chaos is associated with disorder, instability, inconsistency, no control and so on, a common phenomenon in the nature. Using a formal definition, chaos is the irregular movement observed in non linear deterministic dynamical systems 
From the chaos perspective, many temporal series are understood and analyzed. There is a increasing interest to predict temporal series because they can represent real phenomenons. By consequence, many methods based on kernel as Support Vector Machine\cite{Article_16}, Relevant Vector Machine (RVM)\cite{Article_17} and neural network models: Multilayer perceptron\cite{Article_14}, Radial Basis Function Neural Network\cite{Article_15}, Recurrent Predictor Neural Network\cite{Article_20}, Elman Network\cite{Article_21}, and Echo state network (ESN)\cite{Article_9}. Besides, other methods based on Fuzzy Model\cite{Article_18}, Extended Kalman filter\cite{Article_19}.

Recurrent Neural Networks are a powerful tool based on biological brain, there is a great diversity of applications. However, the main difficulty is the training step, because all the weights of the network must be trained\cite{tutorial_Jaeger}.


Reservoir computing (RC) is a recent technique of Machine Learning based on the design of Recurrent Networks, the difference is the simplicity of the training step, only a the weights of output layer is trained. Echo State Network Models(ESNs), Liquid State Machines(LSMs), Backpropagation Decorrelation Neural Networks and Evolution of Systems with Linear Outputs(Evolino) are examples of RC. \cite{Jaeger_1}, proposed ESN to find a way of predicting temporal series.


ESN is a kind of RNNS with a features of simplicity and less computational cost. ESN classical models are build using a big number of neurons with sparse and random connections where only weights of output layer are trained. ESN has big potential for non linear approximation problems and his use is big in the prediction of temporal series: Identification Systems\cite{Jaeger_2}, Modeling neural plasticity for classification and regression\cite{Article_22}, Time series Prediction \cite{Article_9}, Pattern Recognition\cite{Article_23}.


The reservoir of a ESN is used as processing layer and this is not modified during training phase. For a good performance of ESN, the reservoir must satisfied a condition about his state dynamics(echo property), the state of reservoir is an echo of the history of his  inputs. Spectral Radius of the reservoir has a high impact over the performance model and his capacity of good estimations\cite{tutorial_Jaeger}.

Complex Networks are added to the design of the reservoir dynamics of original model of ESN to imitate the way how learning mechanisms of real world works, like Small World and Free Scale Complex Networks\cite{Article_5}. Giving ESN the potential of solving non linear modelling problems and prediction of chaotic temporal series, showing similar features of biological neural networks, like a power-law distribution, small-word feature, community structure, and distributed architecture\cite{Article_8}.


Small World is a feature in a Complex Network with high degree of clustering with a property of a high number of connections between closest nodes forming a small world. In contrast, Free Scale, the connection of nodes is represented by Potency Law. Recently, Neural Networks with Complex Network Topology presented better performance than classical ones\cite{Article_12,Article_13}.



The present work is organized: in Section II, a literature review of ESN and extentions are presented. In Section III, a briefly description ESN is summarized, in Section IV the results of experiments are presented. Finally, Section V states with the conclusions. 


\section{LITERATURE REVIEW}

The classic ESN model has recently been criticized\cite{Problemas_esn_Jaeger} for the fact that the reservoir is generated randomly, as well as the connections of the internal units. This fact causes high complexity in the performance on the model and in many cases does not guarantee the stability of the prediction. This is because the classic ESN model has difficulties due to the lack of information about the understanding and the properties of the reservoir, such as connection configuration and weight allocation, it does not guarantee an optimal configuration. In addition to, it does not providing a clear view of the design and structure about model. That's the reason which motivated many researchers to carry out studies in order to find those parameters that optimize and guarantee the stability of the solution.


As a solution to the randomness' problem, several studies have been developed, finding two strong approaches: modification of weights \cite {Article_11} and modification of the reservoir's structure\cite{Article_8}. On the other hand, Multiple Loops Reservoir structure (MLR) model\cite{Article_3} was proposed and compared to the Adjacent-feedback loop Resevoir (ALR) model, MLR strengthens the connections within the reservoir and improves the skills of the classic model for the nonlinear approach. The work also analyzes the influence of the parameters of the proposed model based in the prediction accuracy. Attending to the same goal, in 2014, a new ESN model known as Simple Cycle Reservoir Network (SCRN)\cite {Article_4} was introduced, where the reservoir was constructed deterministically. Initially the proposal considered a reservoir of size larger than required and using a pruning algorithm, in this case, the Sensitive Iterated Pruning Algorithm algorithm (SIPA) then the less sensitive neurons are turned off to optimize the number of required neurons to reach a good approximation. In the work \cite {Article_5}, a design of clustered complex echo state network is formulated for the forecast of mobile communications traffic using the Fourier spectrum as prior knowledge to generate the functional clusters.\\


The performance of an ESN network is closely related to its learning paradigm, that is, the training algorithms used, which can be categorized in two ways: batch learning and online sequential learning. In batch learning, the constant system weights are fixed while calculating the error associated with each sample in the input. Whereas the online sequential learning weights are constantly updated, and error (and, therefore, the gradient estimate), so that different weights are used for each input sample. In the online sequential learning scheme the data is processed sequentially as well as the network topology is adapted. RNNs have been widely used following this type of approach. \\


Recently, in \cite{Article_1}, the OSESN-RLS (Online Sequential ESN with Sparse Recursive Least Squares) algorithm was proposed, which controls the size of the network using the penalty restrictions of the $l_0$ and $l_1$ norms into the error criterion. \\


In addition to the learning paradigm, the effectiveness of this type of networks is strongly influenced by the size of the reservoir. According to \cite{tutorial_Jaeger}, the fact that the network is generated randomly means that a greater amount of neurons is needed, because a memory capacity in an network with dynamic reservoir with $N$ nodes is always lower than of amount of reservoir's units  in comparison to traditional neural networks, being one of the possible causes of overfitting. As an alternative solution to this problem, evolutionary algorithms have been used \cite{Article_24, Article_25}  to find an optimal number of neurons for the reservoir, the disadvantage is the computational cost, experience and skill abilities of those who use them. Recently, in \cite{Article_2} Leaky-ESN was proposed, modifying the reservoir units of a classic ESN to replace them with leaky integrator units, facilitating learning of slow dynamics\cite{Article_27}. This method considers the output feedback line in the network, thus maintaining the natural essence of a recurring network that generally ignores this type of connections. In addition to, it modifies the sufficient conditions for the echo condition and uses the barrier method to optimize the control of the network parameters.

\section{ESN MODEL}                                     
The Echo state network (ESN) neural network model is a new and robust type of recurring network that has a special configuration. It consists of a network whose hidden layer is a reservoir with a certain number of neurons, where only the weights of the network are trained during the learning phase. For this model to work, it is necessary that the dynamics of the reservoir be less chaotic as possible, it is known as `` echo property''. According to Jaeger (2002)\cite{tutorial_Jaeger}, we define the previously assumption in the next definition. 


\begin{mydef}[Echo property]
A network has the echo status property if the current state of the network is only determined by the values of the past inputs and outputs. In other words for each internal unit of the $r_i$ reservoir, there is a $e_i$ function, which is known as an echo function, which maps the input/output pair ($x_i, y_i$) of its history to the current state. Thus the current state for one reservoir's unit $r_i(n)$ in the time instance $t$ 
is given by the equation (\ref{eco_equation})
\begin{equation}
r_i(n)=e_i((x(n),y(n)),(x(n-1),y(n-1)),\dots)
\end{equation}
\label{eco_equation}
\end{mydef}

\begin{figure}
  \centering
    \includegraphics[width=0.45\textwidth]{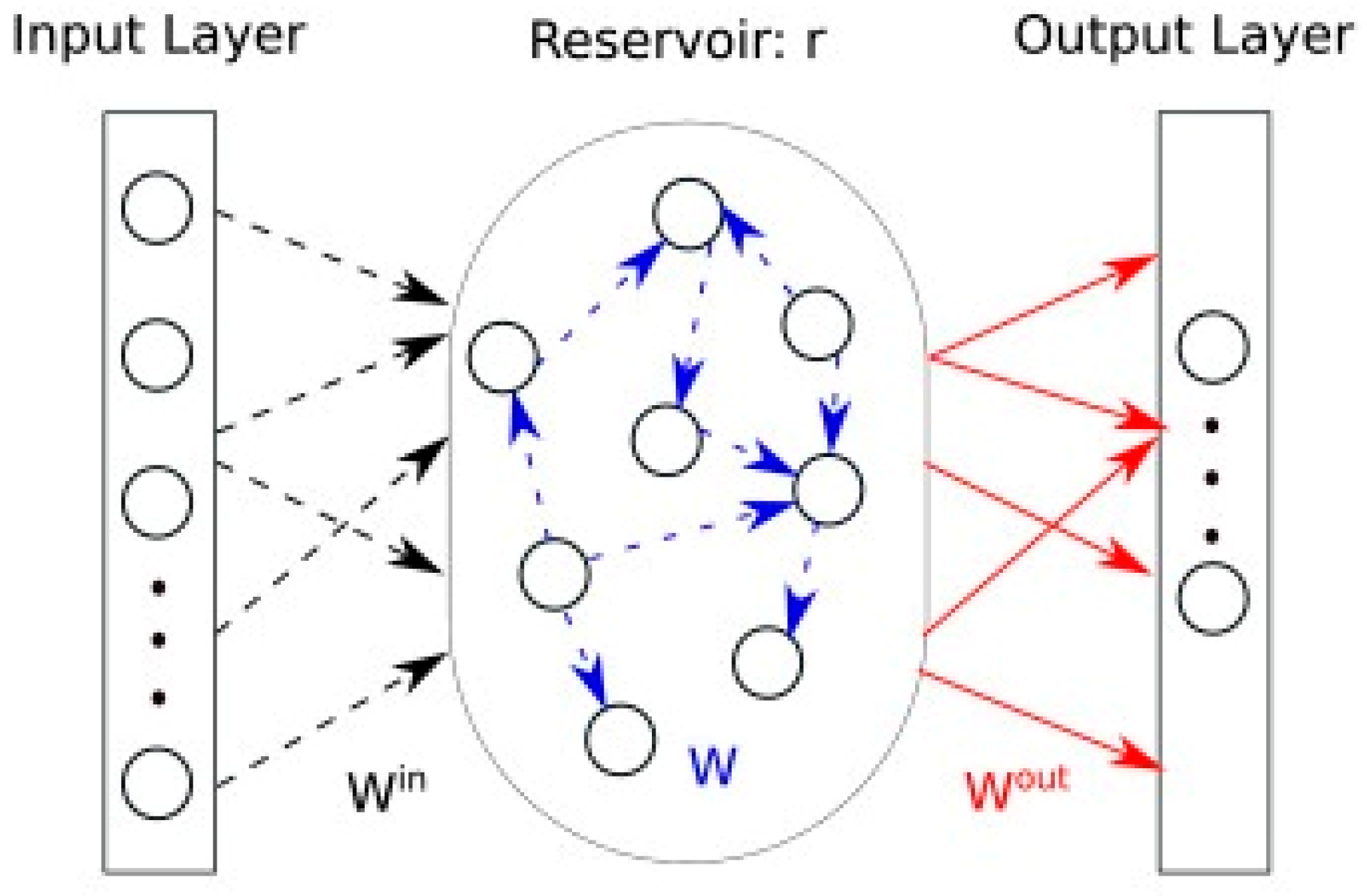}
  \caption{Basic setup of ESN\cite{Article_26}.}
  \label{fig:modelo_esn}
\end{figure}

We are going to consider in the case of time a RNN that has $K$ input units $\mathbf{x} = (x_1, x_2, x_3, \dots, x_K)$, a hidden layer consisting of a reservoir with N internal units, $\mathbf{r} = (r_1, r_2, r_3, \dots, r_N)$, where $\mathbf{r}$ represents the system state vector for some time of time $t$, and a output layer with $L$ units, $\mathbf{y} = (y_1, y_2, y_3, \dots, y_L)$. The weights of the connections between the neurons are stored in the adjacency matrices $ W_{in}, W, W_{out}, W_{back}$ (the last case matrix only if exists feedback), where $W$ is a matrix which stored the weights of the resevoir's units. The sizes of these matrices are $N \times K, N \times N, L \times (K + N + L), N \times L$, respectively. The figure \ref{fig:modelo_esn} shows an example of the architecture of a ESN classical model.


For the implementation of the model we follow the design and ideas proposed in \cite{Original_Article}. Therefore, the activation function for the internal units of the reservoir is given by the equation
$$\mathbf{r}(n+1)=(1-\alpha)\mathbf{r}(t)+\alpha f(W\mathbf{r}(t)+W_{in}\mathbf{x}(t)+\zeta) $$
where $0<\alpha \leq 1$ is `` the leakage rate" which causes the reservoir to evolve more slowly as $\alpha \longrightarrow 0$, $\zeta$ bias and the function $f = tanh$. On the other hand, the output of the network is given by the equation
$$\mathbf{y}(t) = W_{out} \mathbf{r}(t) + c$$
The adjacency matrix of the reservoir, $W$, is obtained through the different configurations of the complex network designs used in the experiments. According to Jaeger(2001)\cite{tutorial_Jaeger}, the echo property for a neural network is closely related to the algebraic properties of the reservoir adjacency matrix, so that a condition necessary for the echo property to be guaranteed is that said matrix has a spectral radius of less than 1. So in this case, we scale the $ W $ matrix so that we guarantee that the model works.


\section{EXPERIMENTS AND RESULTS}




\subsection{Problem}

Let be a dynamical system: $d \phi/dt = f(\phi)$ and input, output vector of the system:  $\mathbf{x}(t),\mathbf{y}(t)$. Where $\phi$ represents the state of the system.  In this problem, $\mathbf{x} \in \mathbb{R}^{N}$ and $\mathbf{y} \in \mathbb{R}^{M}$ are known during the interval of time $[0,T]$ para $T \in \mathbb{R}$.\\


\subsection{Purpose}

The main objective is to estimate the values of $\mathbf{y}$ for a time $t>T$, from knowing the value of $\mathbf{x}$, this is known as an `observer"\cite{Original_Article}.
For this task, a technique from Machine Learning is used, ``reservoir computing". This network originally proposed by Jaeger y Haas\cite{Jaeger_1}. A experimental analysis of the parameters were performed and make an adaptation to change the complex network architecture to analyze the influence of the kind of CN to know the performance, benefits and restrictions.

\subsection*{Rossler Atractor Dynamical System}
Let consider the next equation system:
\begin{equation}
\begin{cases}
dx/dt=-y-z,\\
dy/dt=x+ay,\\
dz/dt=b+z(x-c)
\end{cases}
\label{eq:Rosslersystem}
\end{equation}

then the study is on the Equation Eq. \ref{eq:Rosslersystem} for the instance of $a=0.5$, $b=2.0$ y $c=4.0$. For the experiment, the known and available variable is $x$ so $x$ is the input with the information of the variables $y$ y $z$ together in the interval of time $[T_0,T_1]$, for $T_0>0$. The data used in this experiment was obtained by solving the equation's system described in (\ref{eq:Rosslersystem}) using Runge Kutta 4th order method, with $0.1$ time step.
 
\begin{align*}
     \textrm{number of reservoir nodes: } & N=400 \\
     \textrm{spectral radius: } & \rho=1\\
     \textrm{average degree: } & D=20\\
     \textrm{bias constant: } & \zeta=1\\
     \textrm{leakage rate: } & \alpha=1\\
     \textrm{time step: } & \textrm{delta }t=0.1\\
     \textrm{initial time: } & T_0=100\\
     \textrm{initial time of training phase: } & T_1=260\\
     \textrm{final time of predicted phase: } & T_2=500
\end{align*}

The values of the tested parameters follows the next criterion:
$List\_X$ = A:inc:B, the list starts with $A$ value to $B$ and an increment of $inc$.

\subsection{Number of Reservoir Nodes}

\begin{figure}[!hbt]
  \centering
    \includegraphics[width=0.4\textwidth]{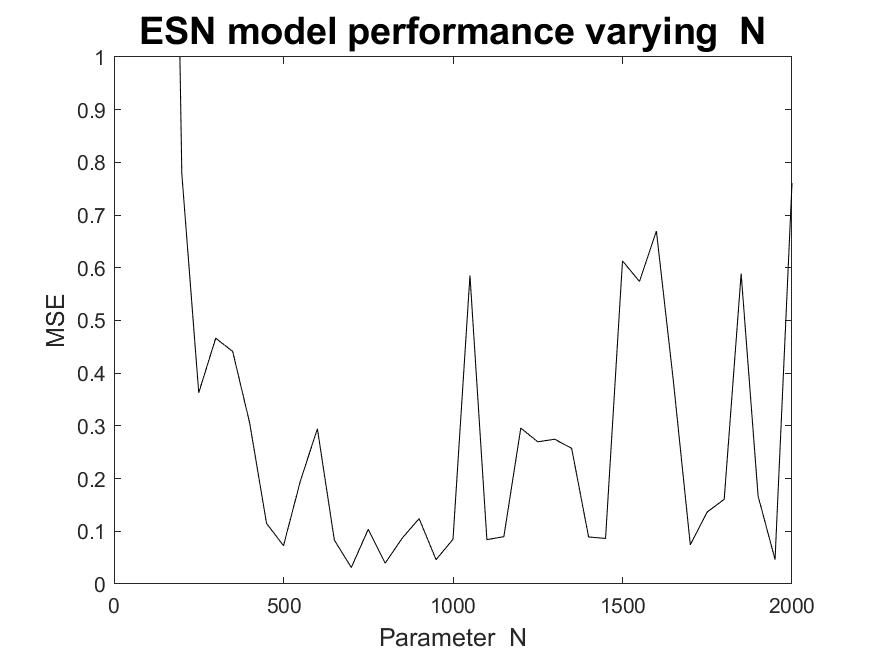}
  \caption{ESN model performance using Erdos Renyi, varying the number of neurons into the reservoir} 
  \label{fig:ER_Parameter_N_Rossler}
\end{figure}

The values of N tested:
$List_N$ = 50:50:2000. The Figure Fig. \ref{fig:ER_Parameter_N_Rossler} presents the results of varying the number of reservoir nodes.

The best results are around the interval [$600$, $900$] neurons.

\subsection{Mean Degree of The Connections}

\begin{figure}[!hbt]
  \centering
    \includegraphics[width=0.4\textwidth]{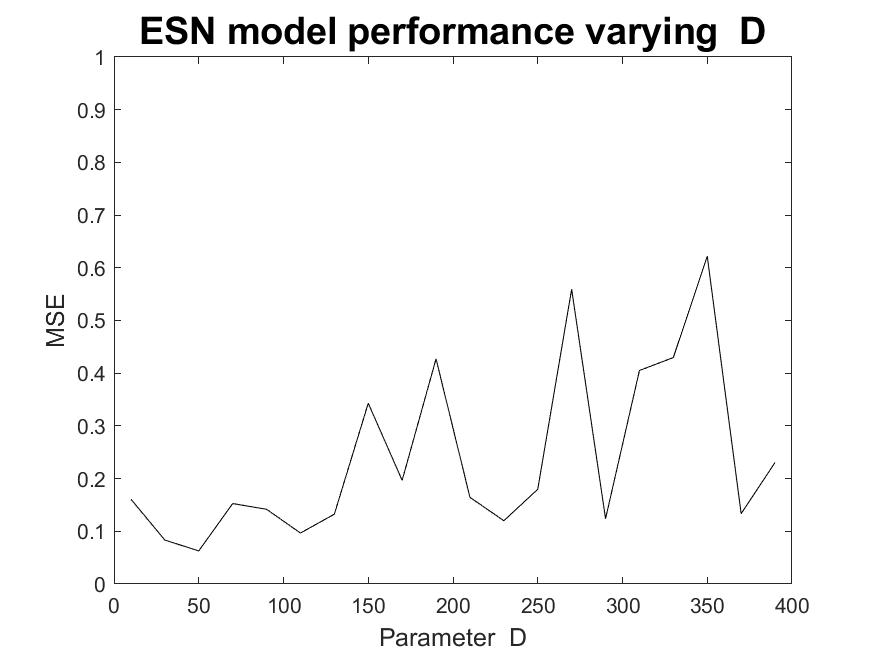}
  \caption{ESN model performance using Erdos Renyi, varying the mean degree into the reservoir} 
  \label{fig:ER_Parameter_D_Rossler}
\end{figure}

The values of parameter $D$:
$List\_D$=10:20:390. The graphic Fig.\ref{fig:ER_Parameter_D_Rossler} presents the results of different mean degree.

In contrast of the previous parameter N, this parameter presents more specific good values, around $50$.

\subsection{Initial Time for Training Step}

\begin{figure}[!hbt]
  \centering
    \includegraphics[width=0.4\textwidth]{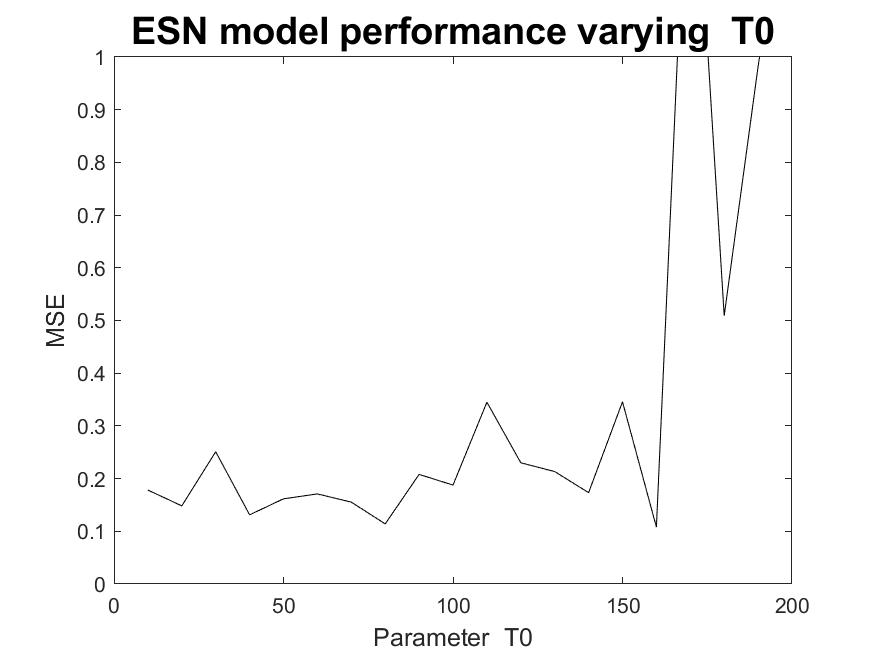}
  \caption{ESN model performance using Erdos Renyi, varying the initial time instance of training phase}
  \label{fig:ER_Parameter_T0_Rossler}
\end{figure}

Values for parameter $T0$, $List\_T0$ = 10:10:200. The plot presented in Fig. \ref{fig:ER_Parameter_T0_Rossler}
 present the MSE values for parameter $T0$.

\subsection{Initial/Final time for Prediction Step}

\begin{figure}[!hbt]
  \centering
    \includegraphics[width=0.4\textwidth]{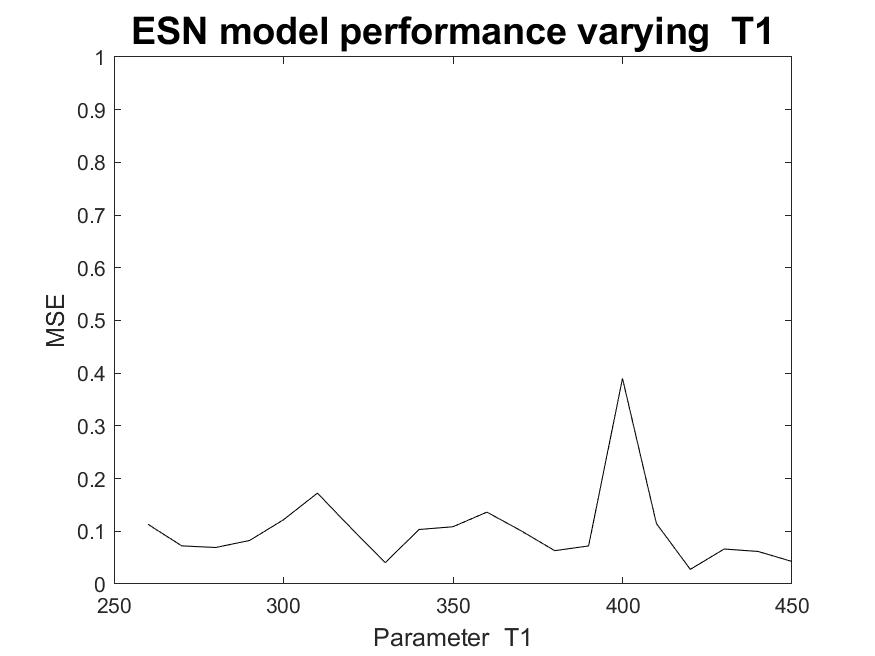}
  \caption{ESN model performance using Erdos Renyi, varying the initial time instance of prediction phase} 
  \label{fig:ER_Parameter_T1_Rossler}
\end{figure}

\begin{figure}[!hbt]
  \centering
    \includegraphics[width=0.4\textwidth]{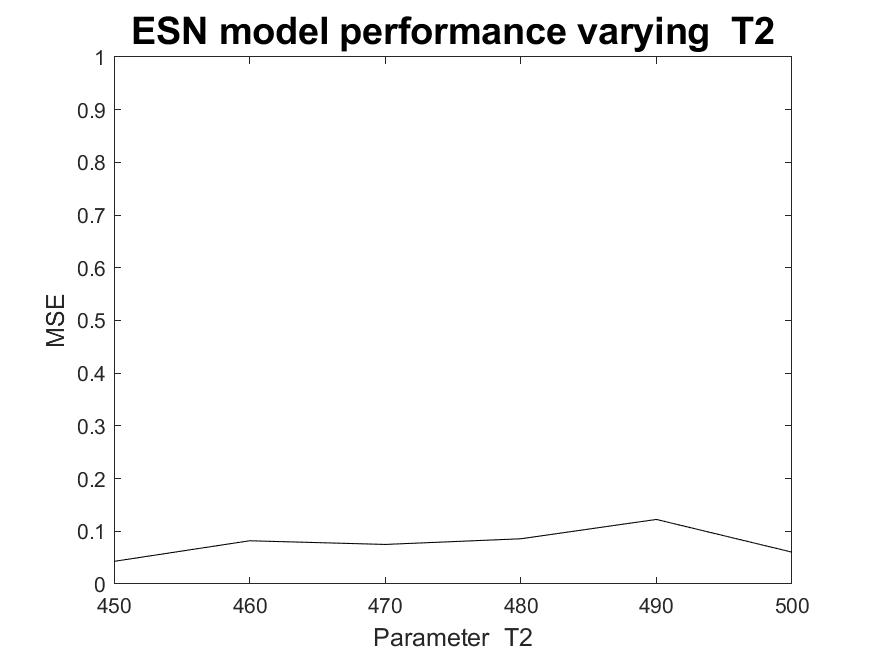}
  \caption{ESN model performance using Erdos Renyi, varying the final time instance for prediction phase} 
  \label{fig:ER_Parameter_T2_Rossler}
\end{figure}

The values for parameters $T1$ and $T2$ are: $List\_T1$ = 260:10:450 and $List\_T2$ = 450:10:800, respectively. The figure Fig. \ref{fig:ER_Parameter_T1_Rossler} and \ref{fig:ER_Parameter_T2_Rossler} presents the results for the experiment.

\subsection{Time Step for Prediction phase}

\begin{figure}[!hbt]
  \centering
    \includegraphics[width=0.4\textwidth]{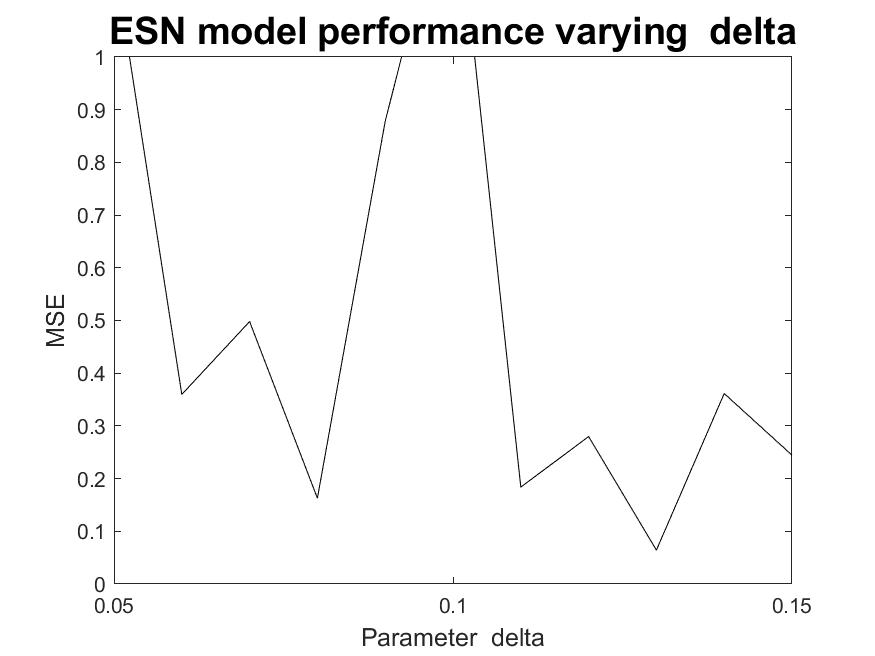}
  \caption{ESN model performance using Erdos Renyi, varying the parameter delta} 
  \label{fig:ER_Parameter_delta_Rossler}
\end{figure}

The graphic Fig. \ref{fig:ER_Parameter_delta_Rossler} shows the values for parameter $delta$.

From the figures: Fig. \ref{fig:ER_Parameter_N_Rossler}, \ref{fig:ER_Parameter_D_Rossler}, \ref{fig:ER_Parameter_T0_Rossler}, \ref{fig:ER_Parameter_T1_Rossler},\ref{fig:ER_Parameter_T2_Rossler}, \ref{fig:ER_Parameter_delta_Rossler}. The first analyzis shows that the parameter $N$ and $D$, $delta$ presents higher inestability or influence with a variation of their values. In contrast, parameter $T0$, $T1$ and $T2$ has high stability or less variance in MSE values.   

\subsection{Complex Network Architecture}

In the table Tab. \ref{tab:table_mse}, a variation of MSE(minimal square error) is presented with the change of Complex Network Architecture for the ESN model\cite{Original_Article}. The Complex Networks tested were: Erd$\ddot{o}$s Renyi, Barabási, Small world and Random Matrix(reservoir randomly generated)

\begin{table}[H]
\centering
  \caption{MSE of ESN model varying the reservoir's configuration}
  \label{Table_models}
 \begin{tabular}{||c c ||} 
 \hline
 Reservoir & MSE  \\ [0.5ex] 
 \hline\hline
 Erd$\ddot{o}$s Renyi & 0.0669 \\ 
 \hline
 Random Matrix & 0.1811 \\
 \hline
 Barabási & 0.0808 \\
 \hline
Small World & 0.1638 \\
 \hline
\end{tabular}
\label{tab:table_mse}
\end{table}

From the table Tab. \ref{tab:table_mse}, the best was Erd$\ddot{o}$s Renyi. However, the differences between the architectures are not high.

\begin{figure*}[!hbt]
  \centering{
    \includegraphics[width=0.43\textwidth]{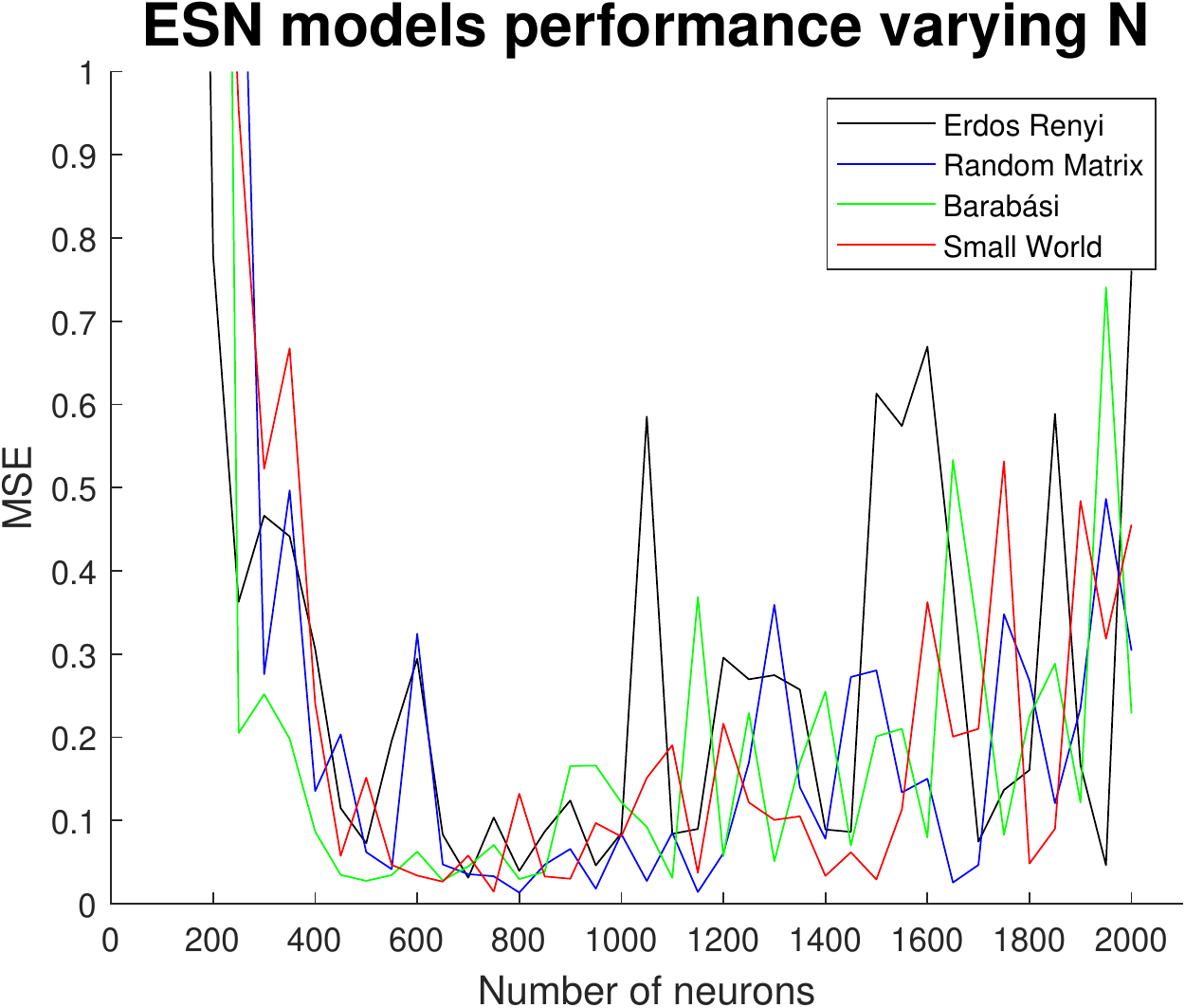}
    \includegraphics[width=0.43\textwidth]{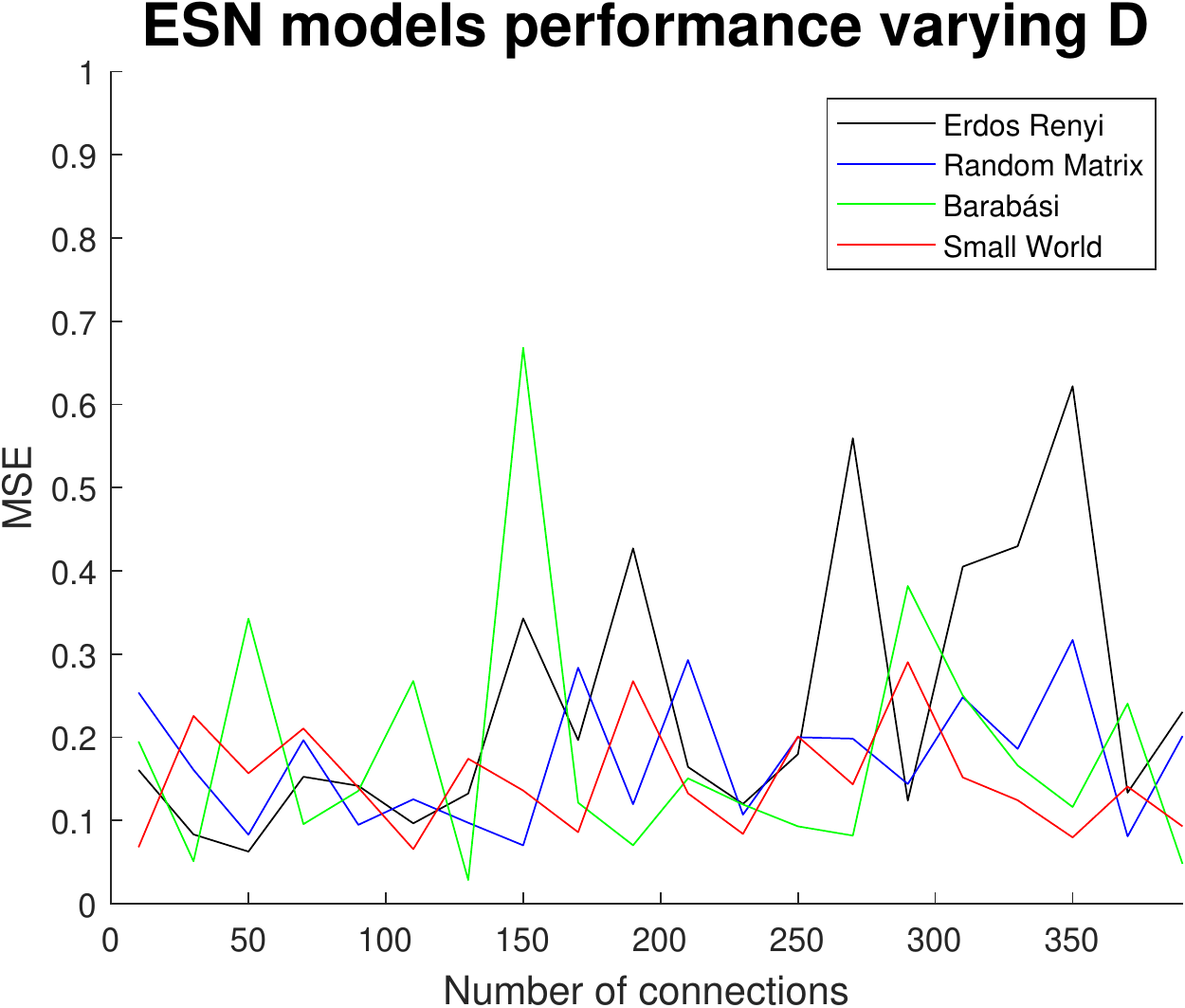}
  }
  \centering{
    \includegraphics[width=0.43\textwidth]{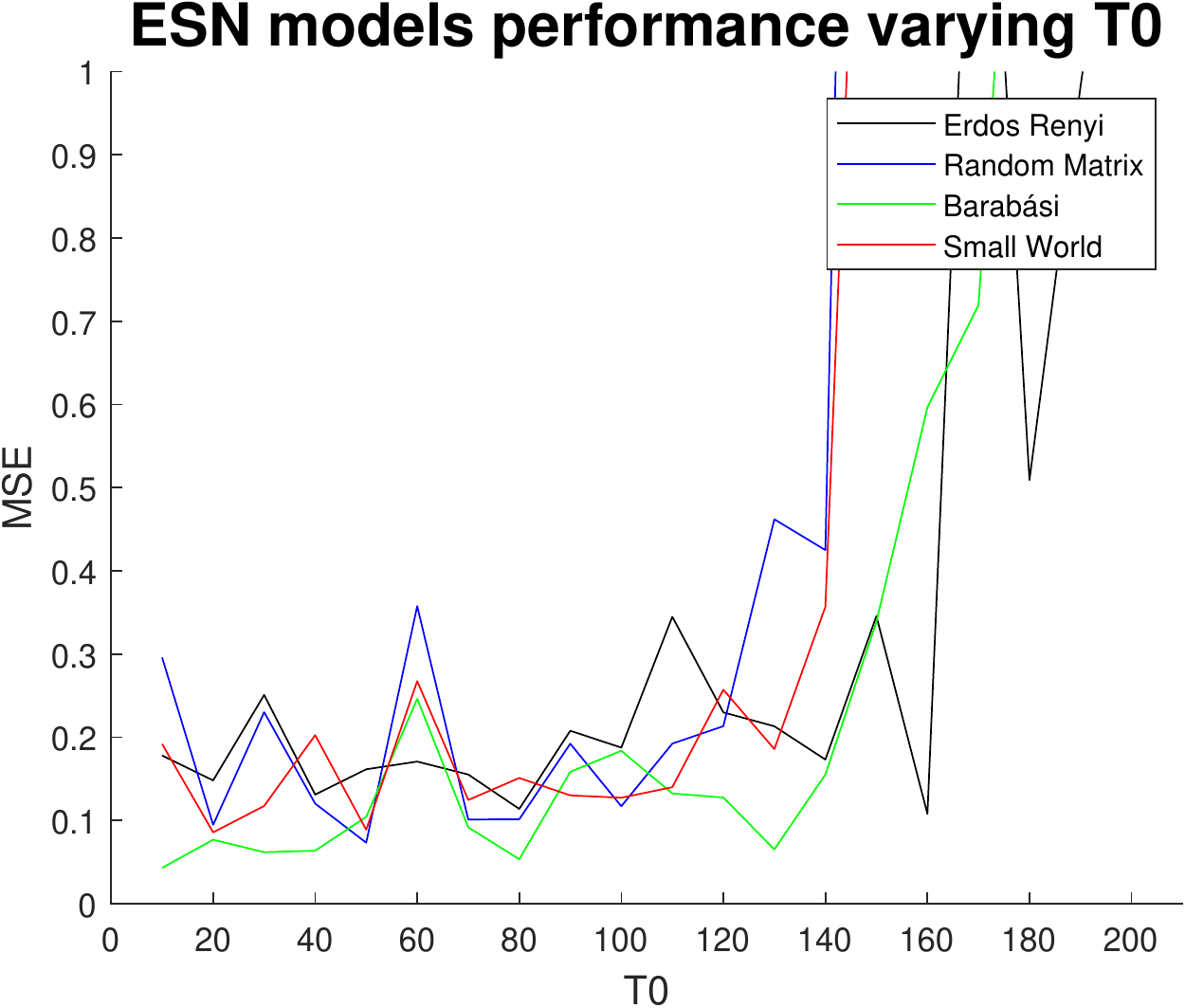}
    \includegraphics[width=0.43\textwidth]{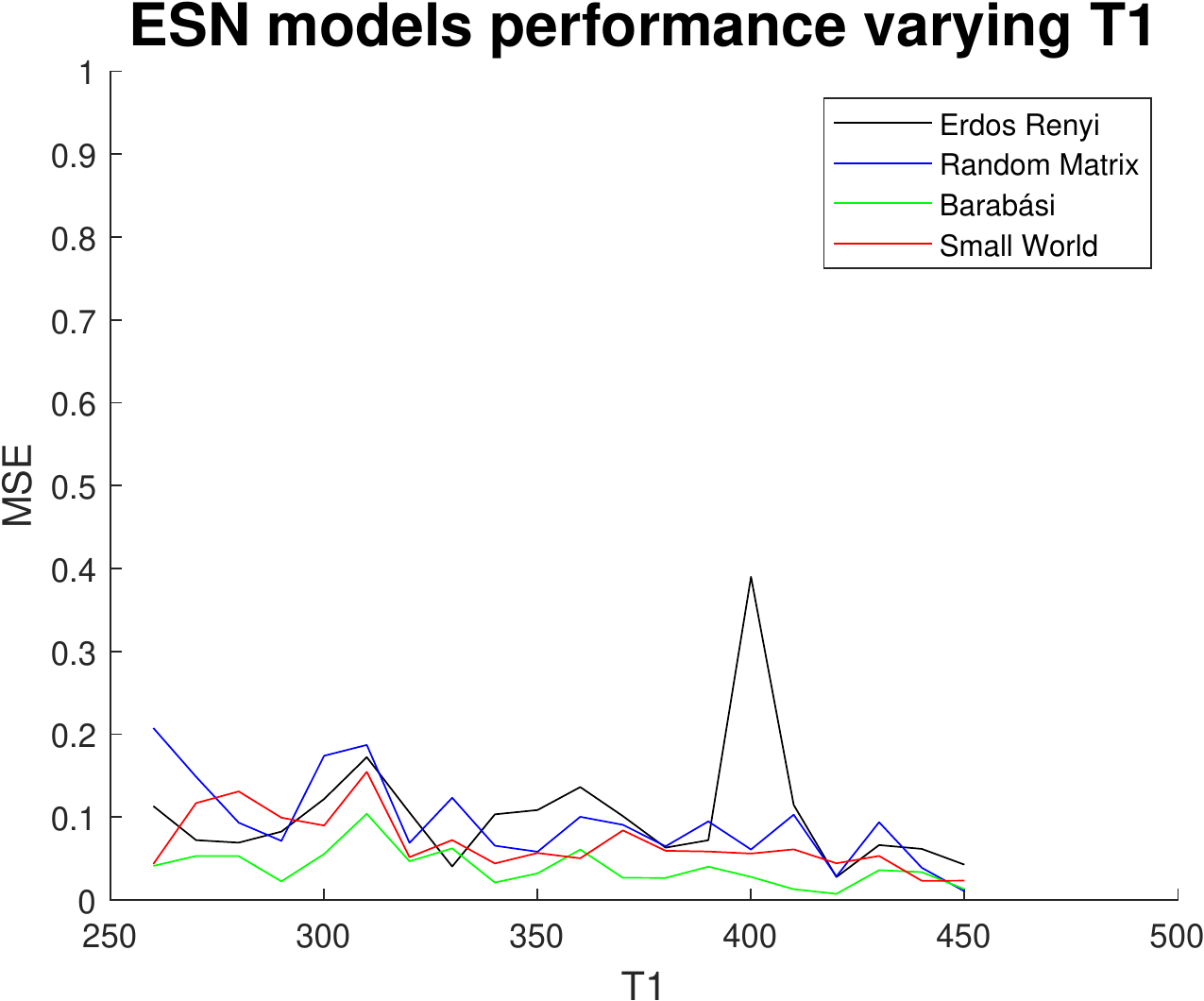}
  }
  \centering{
    \includegraphics[width=0.43\textwidth]{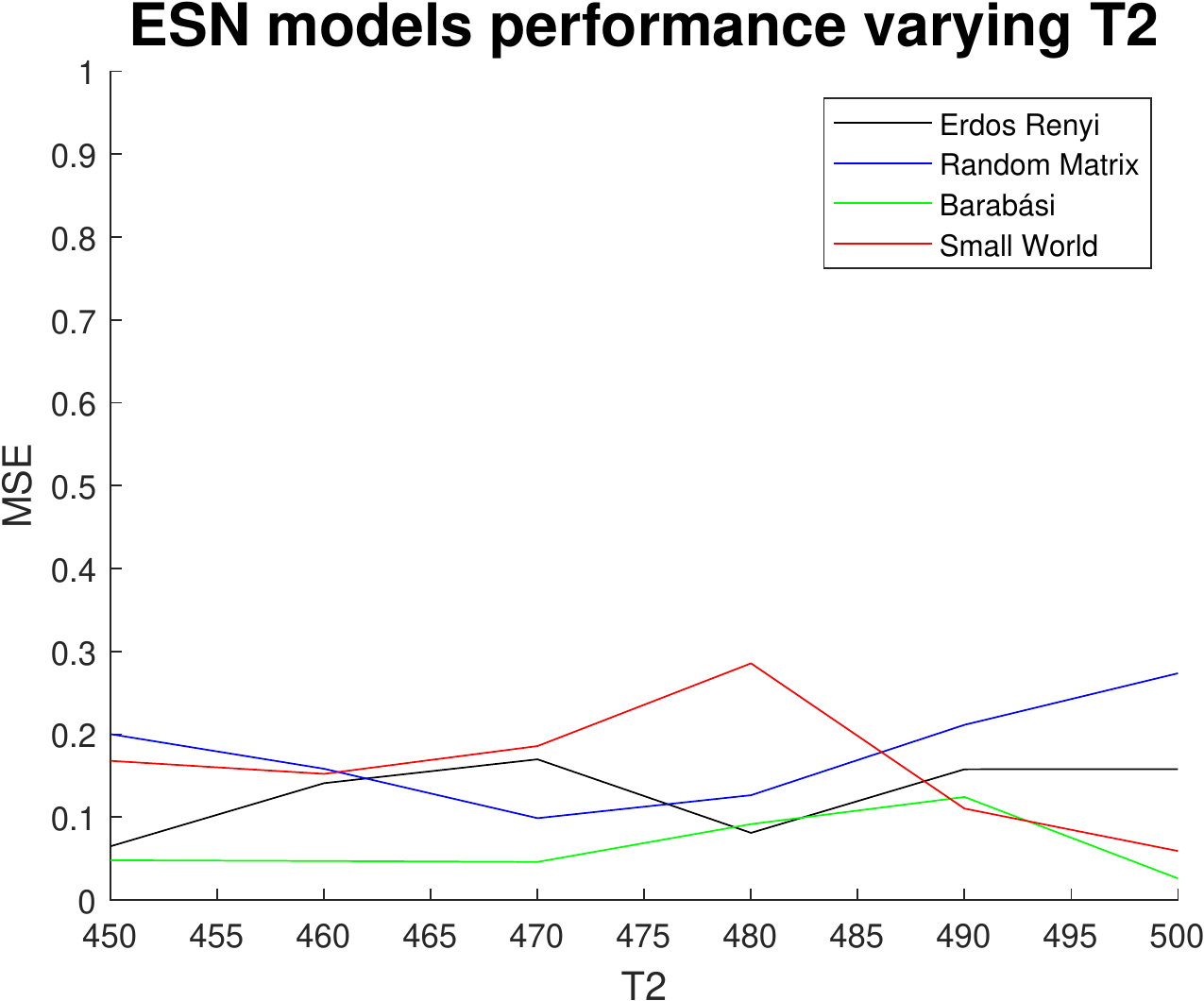}
    \includegraphics[width=0.43\textwidth]{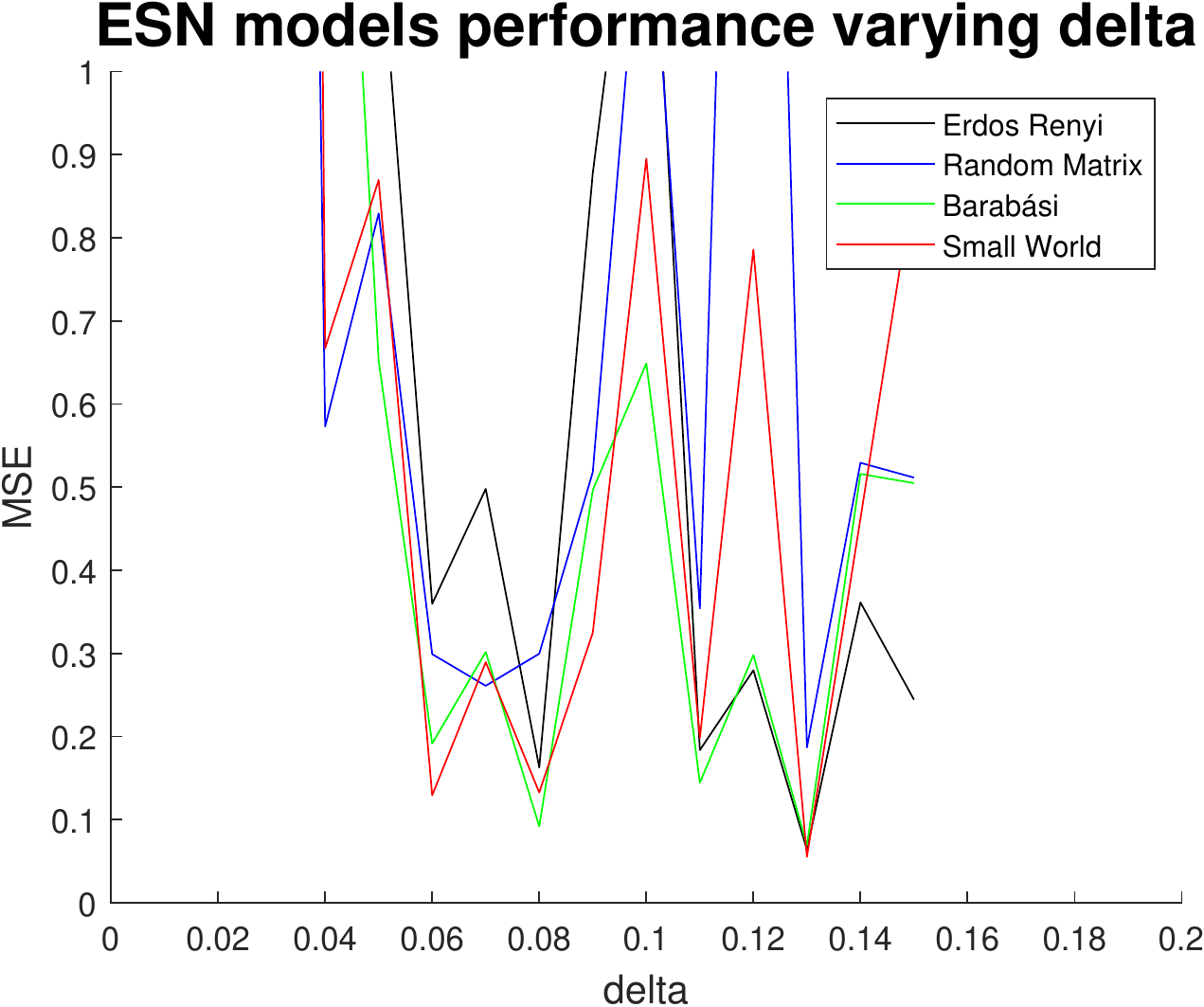}
  }
\caption{Results of Parameter Variation for Different Complex Network Architecture}
\label{fig:Parameter_Exp}
\end{figure*}

The next graphics Fig. \ref{fig:Parameter_Exp} are the result of considering the same criterion of previous experiments using different complex network architecture to analyze visually the impact of the parameters over the performance.

From the results presented in the figure Fig. \ref{fig:Parameter_Exp}, a similar conclusion than previous experiments only using original EsN can be stated. Parameters $N$, $D$ and $delta$ has higher variance in MSE values but a new parameter is found $T0$, the other parameters has less variance. Besides, there is a different behaviour using many kind of Complex Networks and the architecture with similar performance than Erdos Renyi(original ESN model) is Small World even this model overperform Erdos Renyi.



\section{CONCLUSION}
Following the results of the experiments of this work, we can conclude that some parameters studied are potentially influencers than others in the performance for each original ESN model and the performance is very sensible to variety of them. In some cases, a reservoir with a certain configuration(Complex Network architecture) can adjusted with precision to the training data, but at the same time, this fact can generate a deficient generalization and high computational cost, due the reservoirs are constructed based in randomness.



    
\bibliographystyle{IEEEtran}

\bibliography{biblio}

\end{document}